%% file: csn_camera_ready_final.tex
\pdfoutput=1

\documentclass[10pt,twocolumn,letterpaper]{article}

\usepackage{iccv}
\usepackage{times}
\usepackage{graphicx}
\usepackage{amsmath}
\usepackage{amssymb}
\usepackage{tabulary}
\usepackage{multirow}

\usepackage[pagebackref=true,breaklinks=true,letterpaper=true,colorlinks,bookmarks=false]{hyperref}

\iccvfinalcopy 

\def\iccvPaperID{} 

\newcommand{\ddd}[1]{#1$\times$#1$\times$#1}

\ificcvfinal\fi
\begin{document}

\title{Video Classification with Channel-Separated Convolutional Networks}

\author{Du Tran \qquad Heng Wang \qquad Lorenzo Torresani  \qquad Matt Feiszli\\
Facebook AI\\
{\tt\small \{trandu,hengwang,torresani,mdf\}@fb.com}
}

\maketitle

\input{abstract}

\input{intro}

\input{related_work}

\input{3d_group_conv}

\input{experiments}

\input{video_classification}

\input{conclusion}

\input{appendix}

{\small
\bibliographystyle{ieee_fullname}
\bibliography{ieeedu_ref}
}

\end{document}

%% file: abstract.tex
\begin{abstract}

Group convolution has been shown to offer great computational savings in various 2D convolutional architectures for image classification. It is natural to ask: 1) if group convolution can help to alleviate the high computational cost of video classification networks; 2) what factors matter the most in 3D group convolutional networks; and 3) what are good computation/accuracy trade-offs with 3D group convolutional networks.

This paper studies the effects of different design choices in 3D group convolutional networks for video classification. We empirically demonstrate that the amount of channel interactions plays an important role in the accuracy of 3D group convolutional networks. Our experiments suggest two main findings. First, it is a good practice to factorize 3D convolutions by separating channel interactions and spatiotemporal interactions as this leads to improved accuracy and lower computational cost. Second, 3D channel-separated convolutions provide a form of regularization, yielding lower training accuracy but higher test accuracy compared to 3D convolutions. These two empirical findings lead us to design an architecture --  Channel-Separated Convolutional Network (CSN) -- which is simple, efficient, yet accurate. On Sports1M, Kinetics, and Something-Something, our CSNs are comparable with or better than the state-of-the-art while being 2-3 times more efficient.

\end{abstract}

%% file: intro.tex
\section{Introduction}
\label{sec:intro}

Video classification has witnessed much good progress in recent years. Most of the accuracy gains have come from the introduction of new powerful architectures~\cite{I3D,Tran18,P3D,xie2017rethinking,XiaolongWang18}. However, many of these architectures are built on expensive 3D spatiotemporal convolutions.  Furthermore,  these convolutions are typically computed across all the channels in each layer. 3D CNNs have complexity $\mathcal{O}(CTHW)$ as opposed to the cost of $\mathcal{O}(CHW)$ of 2D CNNs, where $T$ denotes the number of frames, $H, W$ the spatial dimensions and $C$ the number of channels.  For both foundational and practical reasons, it is natural to ask which parameters in these large 4D kernels matter the most.

Kernel factorizations have been applied in several settings to reduce compute and improve accuracy.  For example, several recent video architectures factor 3D convolution in space and time: examples include P3D~\cite{P3D}, R(2+1)D~\cite{Tran18}, and S3D~\cite{xie2017rethinking}.  In these architectures, a 3D convolution is replaced with a 2D convolution (in space) followed by a 1D convolution (in time). This factorization can be leveraged to increase accuracy and/or to reduce computation. In the still-image domain, separable convolution~\cite{Xception} is used to factorize the convolution of 2D $k \times k$ filters into a pointwise $1 \times 1$ convolution followed by a depthwise $k \times k$ convolution.  When the number of channels is large compared to $k^2$, which is usually the case, this reduces FLOPs by $\sim k^2$ for images. For the case of 3D video kernels, the FLOP reduction is even more dramatic: $\sim k^3$.

Inspired by the accuracy gains and good computational savings demonstrated by 2D separable convolutions in image classification~\cite{Xception,MobileNet,ShuffleNet}, this paper proposes a set of architectures for video classification -- 3D Channel-Separated Networks (CSN) -- in which all convolutional operations are separated into either pointwise 1$\times$1$\times$1 or depthwise 3$\times$3$\times$3 convolutions. Our experiments reveal the importance of {\it channel interaction} in the design of CSNs. In particular, we show that excellent accuracy/computational cost balances can be obtained with CSNs by leveraging channel separation to reduce FLOPs and parameters as long as high values of channel interaction are retained.  We propose two factorizations, which we call {\it interaction-reduced} and {\it interaction-preserved}. Compared to 3D CNNs, both our interaction-reduced and interaction-preserved CSNs provide higher accuracy and FLOP savings of about 2.5-3$\times$ {\em when} there is enough channel interaction. We experimentally show that the channel factorization in CSNs acts as a regularizer, leading to a higher training error but better generalization. Finally, we show that our proposed CSNs outperform or are comparable with the current state-of-the art methods on Sports1M, Kinetics, and Something-Something while being 2--3 times faster.

%% file: related_work.tex
\section{Related Work}
\label{sec:related_work}

\noindent{\bf Group convolution}. Group convolution was adopted in AlexNet~\cite{Krizhevsky12} as a way to overcome GPU memory limitations. Depthwise convolution was introduced in MobileNet~\cite{MobileNet} as an attempt to optimize model size and computational cost for mobile applications. Chollet~\cite{Xception} built an extreme version of Inception~\cite{Inception} based on 2D depthwise convolution, named Xception, where the Inception block was redesigned to include multiple separable convolutions. Concurrently, Xie~\emph{et al.} proposed ResNeXt~\cite{SainingXie17} by equipping ResNet~\cite{KaimingHe16} bottleneck blocks with groupwise convolution.  Further architecture improvements have also been made for mobile applications. ShuffleNet~\cite{ShuffleNet} further reduced the computational cost of the bottleneck block with both depthwise and group convolution. MobileNetV2~\cite{MobileNet2} improved MobileNet~\cite{MobileNet} by switching from a VGG-style to a ResNet-style network, and introducing a ``reverted bottleneck'' block. All of these architectures are based on 2D CNNs and are applied to image classification while our work focuses on 3D group CNNs for video classification.

\noindent{\bf Video classification}. In the last few years, video classification has seen a major paradigm shift, which involved moving from hand-designed features~\cite{Laptev03,Piotr05,ActionBank,Wang2013} to deep network approaches that learn features and classify end-to-end~\cite{Tran15,Karpathy14,SimonyanZ14,FeichtenhoferNIPS16,WangXW0LTG16,Wang_Transformation,FeichtenhoferPZ16}. This transformation was enabled by the introduction of large-scale video datasets~\cite{Karpathy14,kinetics} and massively parallel computing hardware, \ie, GPU. Carreira and Zisserman~\cite{I3D} recently proposed to inflate 2D convolutional networks pre-trained on images to 3D for video classification. Wang~\emph{et al.}~\cite{XiaolongWang18} proposed non-local neural networks to capture long-range dependencies in videos. ARTNet~\cite{LiminWang18} decouples spatial and temporal modeling into two parallel branches. Similarly, 3D convolutions can also be decomposed into a Pseudo-3D convolutional block as in P3D~\cite{P3D} or factorized convolutions as in R(2+1)D~\cite{Tran18} or S3D~\cite{xie2017rethinking}. 3D group convolution was also applied to video classification in ResNeXt~\cite{retrace} and Multi-Fiber Networks~\cite{MFNet} (MFNet).

Among previous approaches, our work is most closely related to the following architectures. First, our CSNs are similar to Xception~\cite{Xception} in the idea of using channel-separated convolutions. Xception factorizes 2D convolution in channel and space for object classification, while our CSNs factorize 3D convolution in channel and space-time for action recognition. In addition, Xception uses simple blocks, while our CSNs use bottleneck blocks. The variant ir-CSN of our model shares similarities with ResNeXt~\cite{SainingXie17} and its 3D version~\cite{retrace} in the use of bottleneck block with group/depthwise convolution. The main difference is that ResNext~\cite{SainingXie17,retrace} uses group convolution in its \ddd{3} layers with a fixed group size (\eg, $G=32$), while our ir-CSN uses depthwise convolutions in all \ddd{3} layers which makes our architecture fully channel-separated. As we will show in section~\ref{sec:csn_factorization}, making our network fully channel-separated helps not only to reduce a significant amount of compute, but also to improve model accuracy by better regularization. We emphasize that our contribution includes not only the design of CSN architectures, but also a systematic empirical study of the role of channel interactions in the accuracy of CSNs.

%% file: 3d_group_conv.tex
\section{Channel-Separated Convolutional Networks}
\label{sec:channel_separated_network}

In this section, we discuss the concept of 3D channel-separated networks. Since channel-separated networks use group convolution as their main building block, we first provide some background about group convolution.

\subsection{Background}
\label{sec:background}

\begin{figure}
\begin{center}
   \includegraphics[width=\linewidth]{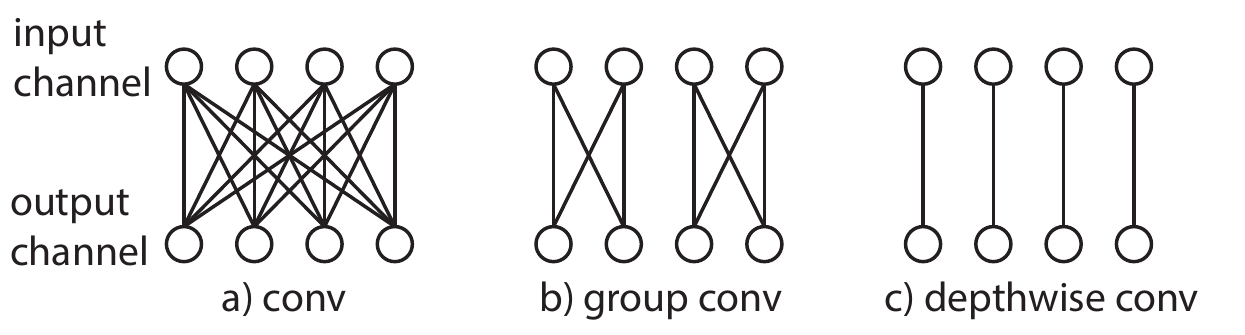}
\end{center}
\vspace{-10pt}
   \caption{{\bf Group convolution}. Convolutional filters can be partitioned into groups with each filter receiving input from channels only within its group. (a) A conventional convolution, which has only one group. (b) A group convolution with 2 groups. (c) A depthwise convolution where the number of groups matches the number of input/output filters, i.e., each group contains only one channel.}
\label{fig:group_conv}
\end{figure}

\noindent{\bf Group convolution}. Conventional convolution is implemented with dense connections, i.e., each convolutional filter receives input from all channels of its previous layer, as in Figure~\ref{fig:group_conv}(a). However, in order to reduce the computational cost and model size, these connections can be sparsified by grouping convolutional filters into subsets. Filters in a subset receive signal from only channels within its group (see Figure~\ref{fig:group_conv}(b)). Depthwise convolution is the extreme version of group convolution where the number of groups is equal to the number of input and output channels (see figure~\ref{fig:group_conv}(c)). Xception~\cite{Xception} and MobileNet~\cite{MobileNet} were among the first networks to use depthwise convolutions. Figure~\ref{fig:group_conv} presents an illustration of conventional, group, and depthwise convolutional layers for the case of $4$ input channels and $4$ output channels.

\noindent{\bf Counting FLOPs, parameters, and interactions}. Dividing a conventional convolutional filter into $G$ groups reduces compute and parameter count by a factor of $G$.  These reductions occur because each filter in a group receives input from only a fraction $1 / G$ of the channels from the previous layer.  In other words, channel grouping restricts feature interaction: only channels within a group can interact. If multiple group convolutional layers are stacked directly on top of each other, this feature segregation is further amplified since each channel becomes a function of small channel-subsets in all preceding layers. So, while group convolution saves compute and parameters, it also reduces feature interactions.

We propose to quantify the amount of channel interaction as the number of pairs of two input channels that are connected through any output filter. If the convolutional layer has $C_{in}$ channels and $G$ groups, each filter is connected to $C_{in}/G$ input channels. Therefore each filter will have ${\frac{C_{in}}{G} \choose 2}$ interacting feature pairs. According to this definition, the example convolutions in Figure~\ref{fig:group_conv}(a)-(c) will have $24$, $4$, and $0$ channel interaction pairs, respectively.

Consider a 3D convolutional layer with spatiotemporal convolutional filters of size \ddd{$k$} and $G$ groups, $C_{in}$ input channels, and $C_{out}$ output channels. Let $THW$  be the total number of voxels in the spatiotemporal tensor provided as input to the layer. Then, the number of parameters, FLOPs (floating-point operations), and number of channel interactions can be measured as:
\begin{eqnarray}
\# \mathrm{parameters} &=& C_{out} \cdot \frac{C_{in}}{G} \cdot k^3 \\
\# \mathrm{FLOPs} &=& C_{out} \cdot \frac{C_{in}}{G} \cdot k^3 \cdot THW \\
\# \mathrm{interactions} &=& C_{out} \cdot {\frac{C_{in}}{G} \choose 2} 
\end{eqnarray}

Recall that ${n \choose 2} = \frac{n(n-1)}{2} = \mathcal{O}\left( n^2 \right)$. We note that while FLOPs and number of parameters are commonly used to characterize a layer, the ``amount'' of channel interaction is typically overlooked. Our study will reveal the importance of this factor.

\subsection{Channel Separation} 
\label{sec:channel_separated_networks}

We define channel-separated convolutional networks (CSN) as 3D CNNs in which all convolutional layers (except for \texttt{conv1}) are either \ddd{1} conventional convolutions or \ddd{$k$} depthwise convolutions (where, typically, $k=3$). Conventional convolutional networks model channel interactions and local interactions (i.e., spatial or spatiotemporal) jointly in their 3D  convolutions. Instead, channel-separated networks decompose these two types of interactions into two distinct layers: \ddd{1} conventional convolutions for channel interaction (but no local interaction) and \ddd{$k$} depthwise convolutions for local spatiotemporal interactions (but not channel interaction). Channel separation may be applied to any \ddd{$k$} traditional convolution by decomposing it into a \ddd{1} convolution and a depthwise \ddd{$k$} convolution.

We introduce the term ``channel-separated'' to highlight the importance of channel interaction; we also point out that the existing term ``depth-separable'' is only a good description when applied to tensors with two spatial dimensions and one channel dimension. We note that channel-separated networks have been proposed in Xception~\cite{Xception} and MobileNet~\cite{MobileNet} for image classification. In video classification, separated convolutions have been used in P3D~\cite{P3D}, R(2+1)D~\cite{Tran18}, and S3D~\cite{xie2017rethinking}, but to decompose 3D convolutions into separate temporal and spatial convolutions. The network architectures presented in this work are designed to separate channel interactions from spatiotemporal interactions.

\subsection{Example: Channel-Separated Bottleneck Block}
\label{sec:lossy_lossless_csn}

\begin{figure}
\begin{center}
   \includegraphics[width=0.8\linewidth]{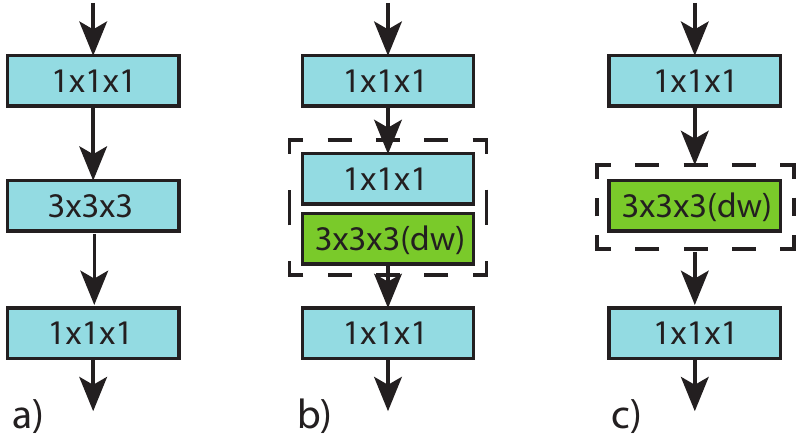}
\end{center}
\vspace{-10pt}
   \caption{{\bf Standard vs. channel-separated convolutional blocks}. (a) A standard ResNet bottleneck block. (b) An interaction-preserved bottleneck block: a bottleneck block where the \ddd{3} convolution in (a) is replaced by a \ddd{1} standard convolution and a \ddd{3} depthwise convolution (shown in dashed box). (c) An interaction-reduced bottleneck block, a bottleneck block where the \ddd{3} convolution in (a) is replaced with a depthwise convolution (shown in dashed box). We note that channel interaction is preserved in (b) by the \ddd{1} convolution, while (c) lost all of the channel interaction in its \ddd{3} convolution after factorization. Batch norm and ReLU are used after each convolution layer. For simplicity, we omit the skip connections.}
\label{fig:factorized_blocks}
\end{figure}

Figure~\ref{fig:factorized_blocks} presents two ways of factorizing a 3D bottleneck block using channel-separated convolutional networks. Figure~\ref{fig:factorized_blocks}(a) presents a standard 3D bottleneck block, while Figure~\ref{fig:factorized_blocks}(b) and~\ref{fig:factorized_blocks}(c) present interaction-preserved and interaction-reduced channel-separated bottleneck blocks, respectively.

\noindent {\bf Interaction-preserved channel-separated bottleneck block} is obtained from the standard bottleneck block (Figure~\ref{fig:factorized_blocks}(a) by replacing the \ddd{3} convolution in (a) with a \ddd{1} traditional convolution and a \ddd{3} depthwise convolution (shown in Figure~\ref{fig:factorized_blocks}(b)). This block reduces parameters and FLOPs of the traditional \ddd{3} convolution significantly, but preserves all channel interactions via a newly-added \ddd{1} convolution. We call this an {\it interaction-preserved} channel-separated bottleneck block and the resulting architecture an {\it interaction-preserved channel-separated network} (ip-CSN).

\noindent {\bf Interaction-reduced channel-separated bottleneck block} is derived from the preserved bottleneck block by removing the extra \ddd{1} convolution.  This yields the depthwise bottleneck block shown in Figure~\ref{fig:factorized_blocks}(c). Note that the initial and final \ddd{1} convolutions (usually interpreted respectively as projecting into a lower-dimensional subspace and then projecting back to the original dimensionality) are now the only mechanism left for channel interactions.  This implies that the complete block shown in (c) has a reduced number of channel interactions compared with those shown in (a) or (b). 
We call this design an {\it interaction-reduced} channel-separated bottleneck block and the resulting architecture an {\it interaction-reduced channel-separated network} (ir-CSN).

\subsection{Channel Interactions in Convolutional Blocks}
\label{sec:group_conv_by_blocks}

The interaction-preserving and interaction-reducing blocks in section~\ref{sec:lossy_lossless_csn} are just two architectures in a large spectrum. In this subsection we present a number of convolutional block designs, obtained by progressively increasing the amount of grouping. The blocks differ in terms of compute cost, parameter count and, more importantly, channel interactions.

\begin{figure}
\begin{center}
   \includegraphics[width=0.8\linewidth]{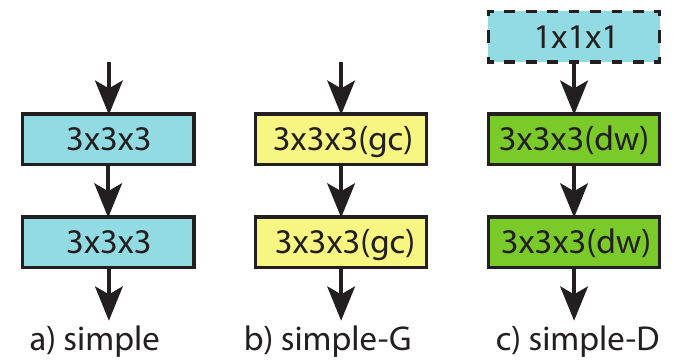}
\end{center}
\vspace{-10pt}
   \caption{{\bf ResNet simple block transformed by group convolution}. (a) Simple block: a standard ResNet simple block with two \ddd{3} convolutional layers. (b) Simple-G block: a ResNet simple block with two \ddd{3} group convolutional layers. (c) Simple-D block: a ResNet simple block with two \ddd{3} depthwise convolutional layers with an optional \ddd{1} convolutional layer (shown in dashed box) added when increasing number of filters is needed. Batch norm and ReLU are used after each convolution layer. For simplicity, we omit the skip connections.}
\label{fig:simpleblock_transform}
\end{figure}

\noindent{\bf Group convolution applied to ResNet blocks}. Figure~\ref{fig:simpleblock_transform}(a) presents a ResNet~\cite{KaimingHe16}  {\bf simple} block consisting of two \ddd{3} convolutional layers. Figure~\ref{fig:simpleblock_transform}(b) shows the {\bf simple-G} block, where the \ddd{3} layers now use grouped convolution.  Likewise, Figure~\ref{fig:simpleblock_transform}(c) presents {\bf simple-D}, with two depthwise layers.  Because depthwise convolution requires the same number of input and output channels, we optionally add a \ddd{1} convolutional layer (shown in the dashed rectangle) in blocks that change the number of channels. 

\begin{figure}
\begin{center}
   \includegraphics[width=\linewidth]{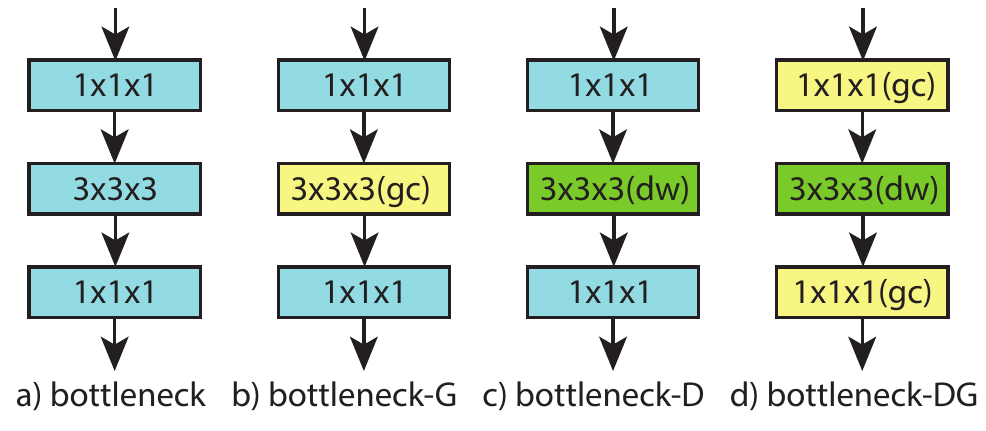}
\end{center}
\vspace{-10pt}
   \caption{{\bf ResNet bottleneck block transformed by group convolution}. (a) A standard ResNet bottleneck block. (b) Bottleneck-G: a ResNet bottleneck block with a \ddd{3} group convolutional layer. (c) Bottleneck-D: a bottleneck block with a \ddd{3} depthwise convolution (previously named as ir-CSN, the new name of Bottleneck-D is used here for simplicity and analogy with other blocks). (d) Bottleneck-DG: a ResNet bottleneck block with a \ddd{3} depthwise convolution and two \ddd{1} group convolutions. We note that from (a) to (d), we gradually apply group convolution to the \ddd{3} convolutional layer and then the two \ddd{1} convolutional layers. Batch norm and ReLU are used after each convolution layer. For simplicity, in the illustration we omit to show skip connections.}
\label{fig:bottleneck_transform}
\end{figure}

Figure~\ref{fig:bottleneck_transform}(a) presents a ResNet {\bf bottleneck} block consisting of two \ddd{1} and one \ddd{3} convolutional layers. Figures~\ref{fig:bottleneck_transform}(b-c) present {\bf bottleneck-G} and  {\bf bottleneck-D} where the \ddd{3} convolutions are grouped and depthwise, respectively. If we further apply group convolution to the two \ddd{1} convolutional layers, the block becomes a {\bf bottleneck-DG}, as illustrated in Figure~\ref{fig:bottleneck_transform}(d). In all cases, the \ddd{3} convolutional layers always have the same number of input and output channels. 

There are some deliberate analogies to existing architectures here.  First, bottleneck-G (Figure~\ref{fig:bottleneck_transform}(b)) is exactly a ResNeXt block~\cite{SainingXie17}, and bottleneck-D is its depthwise variant. Bottleneck-DG (Figure~\ref{fig:bottleneck_transform}(d)) resembles the ShuffleNet block~\cite{ShuffleNet}, without the channel shuffle and without the downsampling projection by average pooling and concatenation. The progression from simple to simple-D is similar to moving from ResNet to Xception (though Xception has many more \ddd{1} convolutions).  We omit certain architecture-specific features in order to better understand the role of grouping and channel interactions.

%% file: experiments.tex
\section{Ablation Experiment}
\label{sec:experiments}

This empirical study will allow us to cast some light on the important factors in the performance of channel-separated network and will lead us to two main findings:
\begin{enumerate}
\item We will empirically demonstrate that within the family of architectures we consider, similar depth and similar amount of channel interaction implies similar accuracy.  In particular, the interaction-preserving blocks reduce compute significantly but preserve channel interactions, with only a slight loss in accuracy for shallow networks and an increase in accuracy for deeper networks.
\item In traditional \ddd{3} convolutions all feature maps interact with each other. For deeper networks, we show that this causes overfitting.
\end{enumerate}

\subsection{Experimental setup}
\label{sec:exp_setup}
\noindent{\bf Dataset}. We use Kinetics-400~\cite{kinetics} for the ablation experiments in this section. Kinetics is a standard benchmark for action recognition in videos. It contains about $260K$ videos of $400$ different human action categories. We use the training split ($240K$ videos) for training and the validation split ($20K$ videos) for evaluating different models.

\noindent{\bf Base architecture}. We use \emph{ResNet3D}, presented in Table~\ref{tab:basic_architecture}, as our base architecture for most of our ablation experiments in this section. More specifically, our model takes clips with a size of T$\times$224$\times$224 where $T=8$ is the number of frames, $224$ is the height and width of the cropped frame. Two spatial downsampling layers (1$\times$2$\times$2) are applied at \texttt{conv1} and at \texttt{pool1}, and three spatiotemporal downsampling (2$\times$2$\times$2) are applied at \texttt{conv3}\_1,  \texttt{conv4}\_1 and  \texttt{conv5}\_1 via convolutional striding. A global spatiotemporal average pooling with kernel size $\frac{T}{8}\times$7$\times$7 is applied to the final convolutional tensor, followed by a fully-connected (fc) layer performing the final classification.

\newcommand{\blocka}[2]{\multirow{3}{*}{\(\left[\begin{array}{c}\text{3$\times$3$ \times$3, #1}\\[-.1em] \text{3$\times$3$ \times$3, #1} \end{array}\right]\)$\times$#2}}

\newcommand{\blockb}[3]{\multirow{3}{*}{\(\left[\begin{array}{c}\text{1$\times$1$ \times$1, #1}\\[-.1em] \text{3$\times$3$ \times$3, #2}\\[-.1em]\text{1$\times$1$ \times$1, #1}\end{array}\right]\)$\times$#3}}

\renewcommand\arraystretch{1.1}
\setlength{\tabcolsep}{3pt}
\begin{table}[t]
\begin{center}
\resizebox{1.0\linewidth}{!}{

\begin{tabular}{c|c|c|c}
\hline
layer name & output size & ResNet3D-simple & ResNet3D-bottleneck \\
\hline
conv1 & T$\times$112$\times$112 & \multicolumn{2}{c}{3$\times$7$\times$7, 64, stride 1$\times$2$\times$2}\\
\hline
pool1 & T$\times$56$\times$56 & \multicolumn{2}{c}{max, 1$\times$3$\times$3, stride 1$\times$2$\times$2}\\
\hline
\multirow{3}{*}{conv2\_x} & \multirow{3}{*}{T$\times$56$\times$56} & \blocka{$64$}{$b_1$} & \blockb{$256$}{$64$}{$b_1$}\\
  &  &  & \\
  &  &  & \\
\hline
\multirow{3}{*}{conv3\_x} &  \multirow{3}{*}{$\frac{T}{2}\times$28$\times$28}  & \blocka{$128$}{$b_2$}  & \blockb{$512$}{$128$}{$b_2$}  \\
  &  &  & \\
  &  &  & \\
\hline
\multirow{3}{*}{conv4\_x} & \multirow{3}{*}{$\frac{T}{4}\times$14$\times$14}  & \blocka{$256$}{$b_3$} & \blockb{$1024$}{$256$}{$b_3$} \\
  &  &  & \\
  &  &  & \\
\hline
\multirow{3}{*}{conv5\_x} & \multirow{3}{*}{$\frac{T}{8}\times$7$\times$7}  & \blocka{$512$}{$b_4$} & \blockb{$2048$}{$512$}{$b_4$} \\
  &  &  & \\
  &  &  & \\
\hline
pool5 & 1$\times$1$\times$1  & \multicolumn{2}{c}{spatiotemporal avg pool, fc layer with softmax} \\
\hline
\end{tabular}
}
\end{center}
\vspace{-.5em}
\caption{{\bf ResNet3D architectures considered in our experiments}. Convolutional residual blocks are shown in brackets, next to the number of times each block is repeated in the stack. The dimensions given for filters and outputs are time, height, and width, in this order. $b_{1,\dots,4}$ are number of blocks implemented at conv2\_x, conv3\_x, conv4\_x, conv5\_x, respectively. The series of convolutions culminates with a global spatiotemporal pooling layer that yields a 512- or 2048-dimensional feature vector. This vector is fed to a fully-connected layer that outputs the class probabilities through a softmax.\vspace{-12pt}}
\label{tab:basic_architecture}
\end{table}

\noindent{\bf Data augmentation}. We use both spatial and temporal jittering for augmentation. Specifically, video frames are scaled such that the shorter edge of the frames becomes $s$ while we maintain the frame original aspect ratio. During training, $s$ is uniformly sampled between $256$ and $320$. Each clip is then generated by randomly cropping windows of size 224$\times$224. Temporal jittering is also applied during training by randomly selecting a starting frame and decoding $T$ frames. For the ablation experiments in this section we train and evaluate models with clips of 8 frames ($T=8$) by skipping every other frame (all videos are pre-processed to 30fps, so the newly-formed clips are effectively at 15fps).

\noindent{\bf Training.} We train our models with synchronous distributed SGD on GPU clusters using caffe2~\cite{caffe2} (with 16 machines, each having $4$ GPUs). We use a mini-batch of $8$ clips per GPU, thus making a total mini-batch of $512$ clips. Following~\cite{Tran18}, we set epoch size to 1M clips due to temporal jitterring augmentation even though the number of training examples is only about 240K. Training is done in $45$ epochs where we use model warming-up~\cite{goyal2017accurate} in the first $10$ epochs and the remaining $35$ epochs will follow the half-cosine period learning rate schedule as in~\cite{slowfast}. The initial learning rate is set to $0.01$ per GPU (equivalent to $0.64$ for 64 GPUs).


\noindent{\bf Testing.} We report clip top-1 accuracy and video top-1 accuracy. For video top-1, we use center crops of $10$ clips uniformly sampled from the video and average these $10$ clip-predictions to obtain the final video prediction.

\subsection{Reducing FLOPs, preserving interactions}
\label{sec:csn_factorization}

In this ablation, we use CSNs to vary both FLOPs and channel interactions.  Within this architectural family, channel interactions are a good predictor of performance, whereas FLOPs are not. In particular, FLOPs can be reduced significantly while preserving interaction count.  

Table~\ref{tab:factorized_models} presents results of our interaction-reduced CSNs (ir-CSNs) and interaction-preserved CSNs (ip-CSNs) and compare them with the ResNet3D baseline using different number of layers. In the shallow network setting (with 26 layers), both the ir-CSN and the ip-CSN have lower accuracy than ResNet3D. The ir-CSN provides a computational savings of 3.6x  but causes a  $2.9\%$ drop in accuracy. The ip-CSN yields a saving of 2.9x in FLOPs with a much smaller drop in accuracy ($0.7\%$).  We note that all of the shallow models have very low count of channel interactions: ResNet3D and ip-CSN have about $0.42$ giga-pairs ($0.42 \times 10^9$ pairs), while ir-CSN has only $0.27$ giga-pairs (about 64\% of the original). This observation suggests that shallow instances of ResNet3D benefit from their extra parameters, but the preservation of channel interactions reduces the gap for ip-CSN.

Conversely, in deeper settings both ir-CSNs and ip-CSNs outperform ResNet3D (by about $0.9-1.4\%$). Furthermore, the accuracy gap between ir-CSN and ip-CSN becomes smaller. We attribute this gap shrinking to the fact that, in the 50-layer and 101-layer configurations, ir-CSN has nearly the same number of channel interactions as ip-CSN since most interactions stem from the \ddd{1} layers.  One may hypothesize that ip-CSN outperforms ResNet3D and ir-CSN because it has more nonlinearities (ReLU). To answer this question, we trained ip-CSNs without ReLUs between the \ddd{1} and the \ddd{3} layers and we observed no notable difference in accuracy. We can conclude that traditional \ddd{3} convolutions contain many parameters which can be removed without an accuracy penalty in the deeper models. We further investigate this next.

We also experimented with a space-time decomposition of the 3D filters~\cite{P3D,Tran18,xie2017rethinking} in ir-CSN-50. This model obtains 69.7\% on Kinetics validation (vs. 70.3\% of vanilla ir-CSN-50) while requiring more memory and having roughly the same GFLOPs as ir-CSN. The small accuracy drop may be due to the fact that CSN 3D filters are already channel-factorized and the space-time decomposition may limit excessively their already constrained modeling ability.

\begin{table}
\begin{center}
{\small 
\begin{tabular}{|c|c|c|c|c|c|}
\hline 
{\bf model} & {\bf depth} & {\bf video@1} & {\bf FLOPs} & {\bf params} & {\footnotesize {\bf interactions}} \\ 
 & & (\%) & $\times 10^9$ & $\times 10^6$ & $\times 10^9$\\
\hline 
ResNet3D & 26 & {\bf 65.3} & 14.3 & 20.4 & 0.42 \\ 
ir-CSN & 26 & 62.4 & 4.0 & 1.7 & 0.27 \\ 
ip-CSN & 26 & 64.6 & 5.0 & 2.4 & 0.42 \\ 
\hline
ResNet3D & 50 & 69.4 & 29.5 & 46.9 & 5.68 \\ 
ir-CSN & 50 & 70.3 & 10.6 & 13.1 & 5.42 \\ 
ip-CSN & 50 & {\bf 70.8} & 11.9 & 14.3 & 5.68 \\ 
\hline 
ResNet3D & 101 & 70.6 & 44.7 & 85.9 & 8.67 \\ 
ir-CSN & 101 & 71.3 & 14.1 & 22.1 & 8.27 \\ 
ip-CSN & 101 & {\bf 71.8} & 15.9 & 24.5 & 8.67 \\ 
\hline
\end{tabular} }
\end{center}
\vspace{-6pt}
\caption{{\bf Channel-Separated Networks vs. ResNet3D}.  In the 26-layer configuration, the accuracy of ir-CSN is $2.9\%$ lower than that of the ResNet3D baseline. But ip-CSN, which preserves channel interactions, is nearly on par with the baseline (the drop is only $0.7\%$). In the the 50- and 101-layer configurations, both ir-CSN and ip-CSN outperform ResNet3D while reducing parameters and FLOPs. ip-CSN consistently outperforms ir-CSN.}
\label{tab:factorized_models}
\vspace{-6pt}
\end{table}

\subsection{What makes CSNs outperform ResNet3D?} 
\begin{figure}
\begin{center}
   \includegraphics[width=0.6\linewidth]{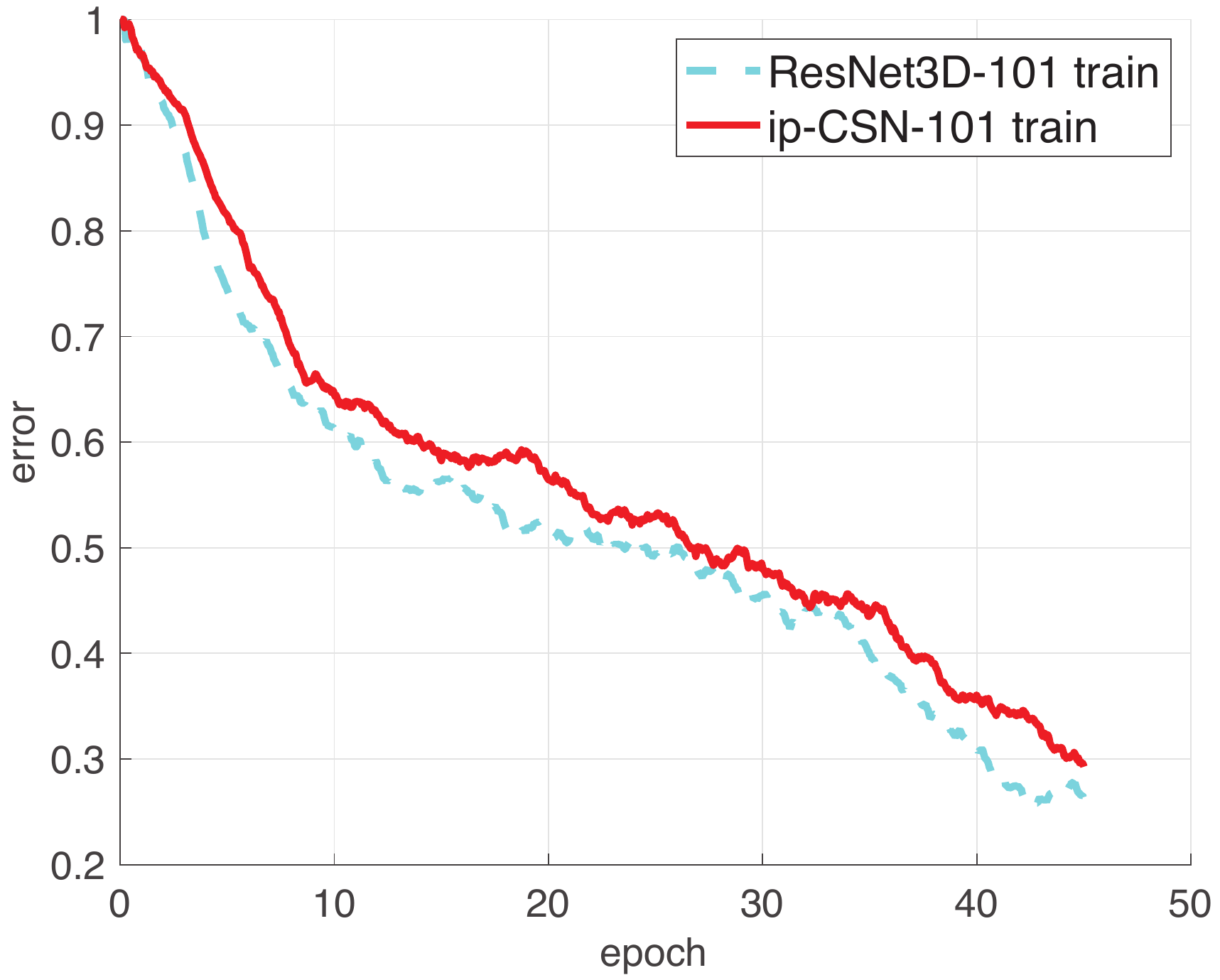}
\end{center}
\vspace{-12pt}
   \caption{{\bf Training error as a function of training iterations for ip-CSN-101 and ResNet3D-101 on Kinetics}. ip-CSN has higher training error, but lower testing error (compare validation accuracies in Table~\ref{tab:factorized_models}). This suggests that the channel separation provides a beneficial regularization, combatting overfitting.}
\label{fig:lossless_csn}
\end{figure}

In section~\ref{sec:csn_factorization} we found that both ir-CSNs and ip-CSNs outperform the ResNet3D baseline when there are enough channel interactions, while having fewer parameters and greatly reducing FLOPs. It is natural to ask: what makes CSNs more accurate? Figure~\ref{fig:lossless_csn} provides a useful insight to answer this question. The plot shows the evolution of the training errors of ip-CSN and ResNet3D, both with 101 layers. Compared to ResNet3D, ip-CSN has higher training errors but lower testing error (see validation accuracy shown in Table~\ref{tab:factorized_models}). This suggests that the channel separation in CSN regularizes the model and prevents overfitting. 

\subsection{The effects of different blocks in group convolutional networks}
\label{sec:block_ablation}

Here we start from our base architecture (shown in Table~\ref{tab:basic_architecture}) then ablatively replace the convolutional blocks with those presented in section~\ref{sec:group_conv_by_blocks}. Again we find that channel interaction plays a critical role in understanding the results.

\noindent {\bf Naming convention}. Since the ablation in this section will be considering several different convolutional blocks, to simplify the presentation, we name each architecture by block type (as presented in section~\ref{sec:group_conv_by_blocks}) and total number of blocks, as shown in the last column of Table~\ref{tab:naming_arch}.

\begin{table}
\begin{center}
{\small 
\begin{tabular}{|c|c|c|c|}
\hline 
{\bf Model} & {\bf block} & {\bf config} & {\bf name} \\ 
\hline
ResNet3D-18 & simple & [2, 2, 2, 2] & simple-8 \\
ResNet3D-50 & bottleneck & [3, 4, 6, 3] & bottleneck-16 \\
\hline 
\end{tabular} }
\end{center}
\vspace{-6pt}
\caption{{\bf Naming convention}. We name architectures by block name followed by the total number of blocks (see last column). Only two block names are given in this table. More blocks are presented in section~\ref{sec:group_conv_by_blocks}.}
\label{tab:naming_arch}
\vspace{-6pt}
\end{table}

\begin{figure*}
\begin{center}
   \includegraphics[width=0.8\linewidth]{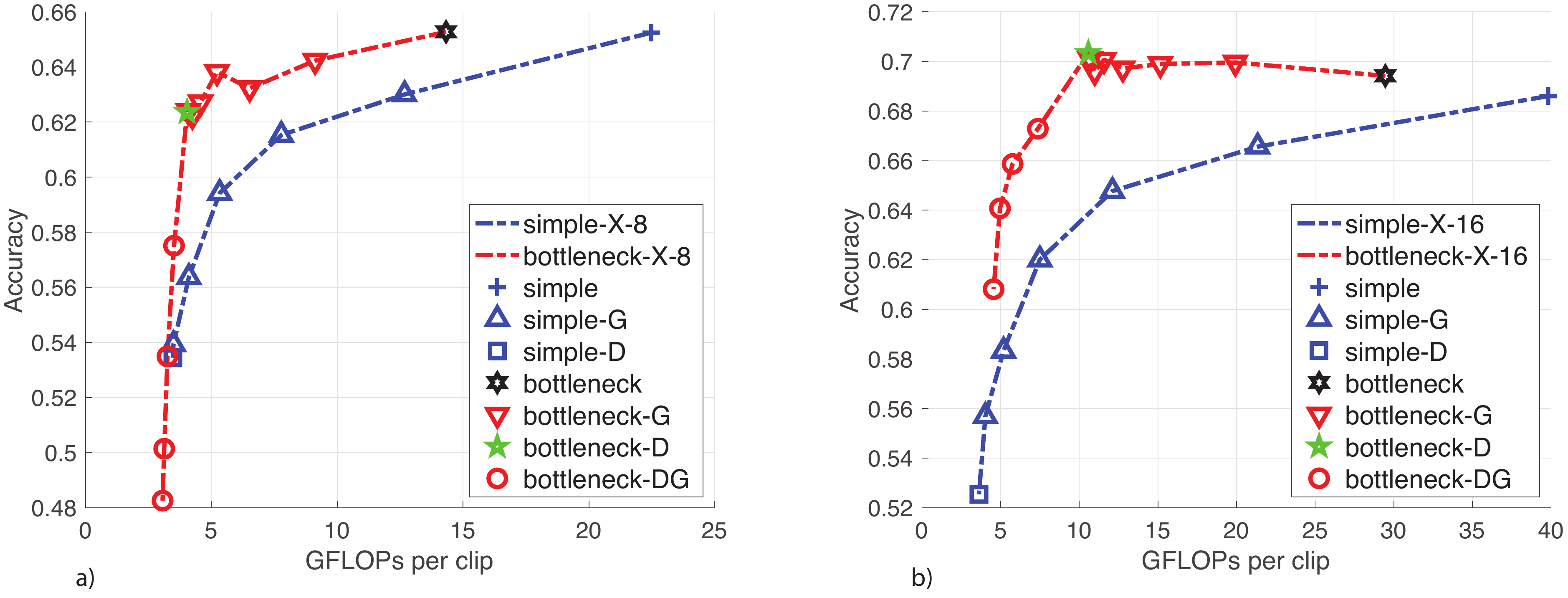}
\end{center}
\vspace{-12pt}
   \caption{{\bf ResNet3D accuracy/computation tradeoff by transforming group convolutional blocks}. Video top-1 accuracy on the Kinetics validation set against computation cost (\# FLOPs) for a ResNet3D with different convolutional block designs. (a) Group convolution transformation applied to simple and bottleneck blocks with shallow architectures with 8 blocks. (b) Group convolution transformation applied to simple and bottleneck blocks with deep architectures with 16 blocks. The bottleneck-D block (marked with green starts) gives the best accuracy tradeoff among the tested block designs. Base architectures are marked with black hexagrams. Best viewed in color.}
\label{fig:block_transform}
\vspace{-6pt}
\end{figure*}

Figure~\ref{fig:block_transform} presents the results of our ablation on convolutional blocks. It shows the video top-1 accuracy on the Kinetics validation set vs the model computational cost (\# FLOPs). We note that, in this experiment, we use our base architecture with two different numbers of blocks (8 and 16) and just vary the type of convolutional block and number of groups to study the tradeoffs. Figure~\ref{fig:block_transform}(a) presents our ablation experiment with simple-X-8 and bottleneck-X-8 architectures (where X can be none, G, or D, or even DG in the case of bottleneck block). Similarly, Figure~\ref{fig:block_transform}(b) presents our ablation experiment with simple-X-16 and bottleneck-X-16 architectures. We can observe the computation/accuracy effects of the group convolution transformation. Reading each curve from right to left (i.e. in decreasing accuracy), we see simple-X transforming from simple block to simple-G (with increasing number of groups), then to simple-D block. For bottleneck-X, reading right to left shows bottleneck block, then transforms to bottleneck-G (with increasing groups), bottleneck-D, then finally to bottleneck-DG (again with increasing groups).  

While the general downward trend is expected as we decrease parameters and FLOPs, the shape of the simple and bottleneck curves is quite different.  The simple-X models degrade smoothly, whereas bottleneck-X stays relatively flat (particularly bottleneck-16, which actually {\it increases} slightly as we decrease FLOPs) before dropping sharply. 

\begin{figure}
\begin{center}
   \includegraphics[width=0.8\linewidth]{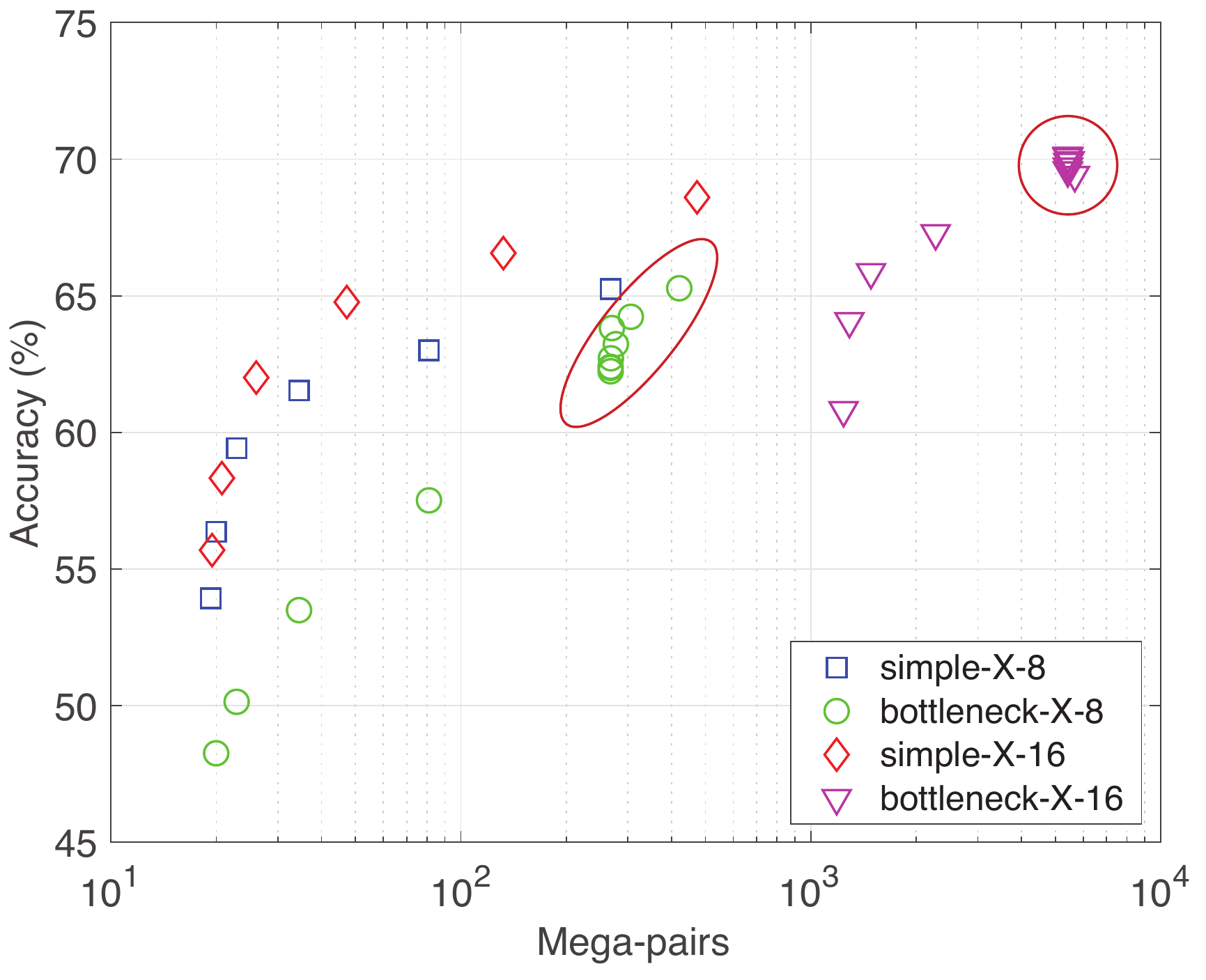}
\end{center}
\vspace{-12pt}
   \caption{{\bf Accuracy vs. channel interactions}. Plotting the Kinetics validation accuracy of different models with respect to their total number of channel interactions. Channel interactions are presented on a log scale for better viewing. Best viewed in color.}
\label{fig:flops_inters_vs_accuracy}
\vspace{-10pt}
\end{figure}

In order to better understand the different behaviors of the simple-X-Y and bottleneck-X-Y models (blue vs. red curves) in Figure~\ref{fig:block_transform} and the reasons behind the turning points of bottleneck-D block (green start markers in Figure~\ref{fig:block_transform}), we plot the performance of all these models according to another view: accuracy vs channel interactions (Figure~\ref{fig:flops_inters_vs_accuracy}). 

As shown in Figure~\ref{fig:flops_inters_vs_accuracy}, the number of channel interactions in simple-X-Y models (blue squares and red diamonds) drops quadratically when group convolution is applied to their 3$\times$3$\times$3 layers. In contrast, the number of channel interactions in bottleneck-X-Y models (green circles and purple triangles) drops marginally when group convolution is applied to their 3$\times$3$\times$3 since they still have many 1$\times$1$\times$1 layers (this can be seen in the presence of two marker clusters which are circled in red: the first cluster includes purple triangles near the top-right corner and the other one includes green circles near the center of the figure). The channel interaction in bottleneck-X-Y starts to drop significantly when group convolution is applied to their 1$\times$1$\times$1 layers, and causes the model sharp drop in accuracy. This fact explains well why there is no turning point in simple-X-Y curves and also why there are turning points in bottleneck-X-Y curves. It also confirms the important role of channel interactions in group convolutional networks.

\noindent {\bf Bottleneck-D block (also known as ir-CSN) provides the best computation/accuracy tradeoff}. For simple blocks, increasing the number of groups causes a continuous drop in accuracy.  However, in the case of the bottleneck block (i.e. bottleneck-X-Y) the accuracy curve remains almost flat as we increase the number of groups until arriving at the bottleneck-D block, at which point the accuracy degrades dramatically when the block is turned into a bottleneck-DG (group convolution applied to 1$\times$1$\times$1 layers).  We conclude that a bottleneck-D block (or ir-CSN) gives the best computation/accuracy tradeoff in this family of ResNet-style blocks, due to its high channel-interaction count.

%% file: video_classification.tex
\section{Comparison with the State-of-the-Art}
\label{sec:video_classification}


\noindent {\bf Datasets}. We evaluate our CSNs on Sports1M~\cite{Karpathy14}, Kinetics-400~\cite{kinetics}, and Something-Something-v1~\cite{GoyalKMMWKHFYMH17}. Sports1M is a large-scale action recognition dataset containing 1.1 million videos from $487$ sport action classes. Kinetics contains about $260K$ videos of $400$ different human action categories. Something$^2$-v1 consists of $100K$ videos of $174$ different object-interaction actions. For Sports1M, we use the public train and test splits provided with the dataset. For Kinetics and Something$^2$-v1, we use the train split for training and the validation set for testing.

\noindent {\bf Training}. Differently from our ablation experiments, here we train our CSNs with $32$-frame clip inputs ($T=32$) with a sampling rate of $2$ (skipping every other frame) following the practice described in~\cite{Tran18}. All the other training settings such as data augmentation and optimization parameters are the same as those described in our previous section. 

\noindent {\bf Testing}. For Sports1M and Something$^2$-v1, we uniformly sample 10 clips per video, scale the shorter edge to 256 (keeping aspect ratio), and use only the center crop of 224$\times$224 per clip for inference. We average the softmax predictions of these 10 crops for video prediction. On Kinetics, since the 30 crops evaluation in~\cite{XiaolongWang18} is widely adopted, we follow this setup for a fair comparison with previous approaches.

\noindent {\bf Results on Sports1M}. Table~\ref{tab:sports1m_results} compares results of our CSNs with those of previous methods on Sports1M. Our ir-CSN-152 and ip-CSN-152 outperform C3D~\cite{Tran15} by $14.4\%$, P3D~\cite{P3D} by $9.1\%$, Conv Pool~\cite{Ng15} by $3.8\%$, and R(2+1)D~\cite{Tran18} by $2.2\%$ on video top-1 accuracy while being 2-4x faster than R(2+1)D. Our ir-CSN-101, even with a smaller number of FLOPs, still outperforms all previous work by good margins. On large-scale benchmarks like Sports1M, the difference between ir-CSN and ip-CSN is very small. The added benefit of ir-CSN is that it has smaller GFLOPs, especially in deeper settings where the number of channel interactions is similar to that of ip-CSN. This is consistent with the observation from our ablation.

\begin{table}
\begin{center}
{\small
\begin{tabular}{|c|c|c|c|c|}
\hline 
{\bf Method} & {\bf input} & {\bf video@1} & {\bf video@5} & {\bf {\scriptsize GFLOPs$\times$crops}} \\ 
\hline 
C3D~\cite{Tran15} & RGB & 61.1& 85.2 & N/A \\ 
P3D~\cite{P3D} & RGB & 66.4 & 87.4 & N/A \\ 
Conv Pool~\cite{Ng15} & {\scriptsize RGB+OF} & 71.7 & 90.4 & N/A \\ 
R(2+1)D~\cite{Tran18} & RGB & 73.0 & 91.5 & 152$\times$N/A \\
R(2+1)D~\cite{Tran18} & {\scriptsize RGB+OF} & 73.3 & 91.9 & 305$\times$N/A \\
\hline
ir-CSN-101 & RGB  & 74.8 & 92.6 & 56.5$\times$10 \\ 
ip-CSN-101 & RGB  & 74.9 & 92.6 & 63.6$\times$10 \\ 
ir-CSN-152 & RGB & {\bf 75.5} & {\bf 92.7} & 74.0$\times$10 \\ 
ip-CSN-152 & RGB  & {\bf 75.5} & {\bf 92.8} & 83.3$\times$10 \\ 
\hline 
\end{tabular} }
\vspace{2pt}
\caption{{\bf Comparison with state-of-the-art architectures on Sports1M}. Our CSNs with 101 or 152 layers outperform all the previous models by good margins while being 2-4x faster. 
}
\label{tab:sports1m_results}
\end{center}
\vspace{-16pt}
\end{table}

\noindent {\bf Results on Kinetics}. We train our CSN models on Kinetics and compare them with current state-of-the-art methods. In addition to training from scratch, we also finetune our CSNs with weights initialized from models pre-trained on Sports1M. For a fair comparison, we compare our CSNs with the methods that use only RGB as input. Table~\ref{tab:kinetics_results} presents the results. Our ip-CSN-152, even when trained from scratch, outperforms all of the previous models, except for SlowFast~\cite{slowfast}. Our ip-CSN-152, pre-trained on Sports1M outperforms I3D~\cite{I3D}, R(2+1)D~\cite{Tran18}, and S3D-G~\cite{xie2017rethinking} by $8.1\%$, $4.9\%$, and $4.5\%$, respectively. It also outperforms recent work: $A^2$-Net~\cite{A2Net} by $4.6\%$, Global-reasoning networks~\cite{global-reasoning} by $3.1\%$. We note that our ip-CSN-152 achieves higher accuracy than both I3D with Non-local Networks (NL)~\cite{XiaolongWang18} and SlowFast~\cite{slowfast} (+$1.5\%$ and +$0.3\%$) while being also faster (3.3x and 2x, respectively). Our ip-CSN-152 is still $0.6\%$ lower than SlowFast augmented with Non-Local Networks. Finally, recent work~\cite{uru_video} has shown that R(2+1)D can achieve strong performance when pre-trained on a large-scale weakly-supervised dataset. We pre-train/finetune ir- and ip-CSN-152 on the same dataset and compare it with R(2+1)D-152 (the last three rows of Table~\ref{tab:kinetics_results}). In this large-scale setup, ip- and ir-CSN-152 outperform R(2+1)D-152 by $1.2$ and $1.3\%$, respectively, in video top-1 accuracy while being $3.0$--$3.4$ times faster.

\begin{table}
\centering
{\small 
 \begin{tabular}{|c|c|c|c|c|}
 \hline
 {\bf Method} & {\bf pretrain} & {\bf vi@1} & {\bf vi@5} & {\bf {\scriptsize GFLOPs$\times$crops}} \\ 
 \hline
ResNeXt~\cite{retrace} & none & 65.1 & 85.7 & N/A \\
ARTNet(d)~\cite{LiminWang18} & none & 69.2 & 88.3 & 24$\times$250 \\ 
I3D~\cite{I3D} & {\scriptsize ImageNet} & 71.1 & 89.3 & 108$\times$N/A \\
TSM~\cite{abs-1811-08383} & {\scriptsize ImageNet} & 72.5& 90.7 & 65$\times$N/A \\
MFNet~\cite{MFNet} & {\scriptsize ImageNet} & 72.8 & 90.4 & 11$\times$N/A \\
{\footnotesize Inception-ResNet}~\cite{abs-1708-03805} & {\scriptsize ImageNet} & 73.0 & 90.9 & N/A \\
R(2+1)D-34~\cite{Tran18} & {\scriptsize Sports1M} & 74.3 & 91.4 & 152$\times$N/A \\
$A^2$-Net~\cite{A2Net} & {\scriptsize ImageNet} & 74.6 & 91.5 & 41$\times$N/A \\
S3D-G~\cite{xie2017rethinking} & {\scriptsize ImageNet} & 74.7 & 93.4 & 71$\times$N/A \\
D3D~\cite{abs-1812-08249} & {\scriptsize ImageNet} & 75.9 & N/A & N/A \\
GloRe~\cite{global-reasoning} & {\scriptsize ImageNet} & 76.1 & N/A & 55$\times$N/A \\
I3D+NL~\cite{XiaolongWang18} & {\scriptsize ImageNet} & 77.7 & 93.3 & 359$\times$30 \\
SlowFast~\cite{slowfast} & none & 78.9 & 93.5 & 213$\times$30 \\
SlowFast+NL~\cite{slowfast} & none & {\bf 79.8} & {\bf 93.9} & 234$\times$30 \\
\hline
ir-CSN-101 & none & 76.2 & 92.2 & 73.8$\times$30 \\ 
ip-CSN-101 & none & 76.7 & 92.3 & 83.0$\times$30 \\ 
ir-CSN-152 & none & 76.8 & 92.5 & 96.7$\times$30 \\ 
ip-CSN-152 & none & 77.8 & 92.8 & 108.8$\times$30 \\ 
\hline
ir-CSN-101 & {\scriptsize Sports1M} & 78.1 & 93.4 & 73.8$\times$30 \\
ip-CSN-101 & {\scriptsize Sports1M} & 78.5 & 93.5 & 83.0$\times$30 \\ 
ir-CSN-152 & {\scriptsize Sports1M} & 79.0 & 93.5 & 96.7$\times$30 \\ 
ip-CSN-152 & {\scriptsize Sports1M} & {\bf 79.2} & {\bf 93.8} & 108.8$\times$30 \\ 
\hline
R(2+1)D-152*~\cite{uru_video} & IG-65M & 81.3 & 95.1 & 329$\times$30 \\ 
ir-CSN-152* & IG-65M & {\bf 82.6} & {\bf 95.3} & 96.7$\times$30 \\ 
ip-CSN-152* & IG-65M & {\bf 82.5} & {\bf 95.3} & 108.8$\times$30 \\ 
 \hline
 \end{tabular}
 }
\vspace{1pt}
\caption{{\bf Comparison with state-of-the-art architectures on Kinetics}. Accuracy is measured on the Kinetics validation set. For fair evaluation, the comparison is restricted to models trained on RGB input. Our ir-CSN-152 is better than or comparable with previous models while being multiple times faster. *Models leveraging large-scale pre-training, thus not comparable with others.}
\label{tab:kinetics_results}
 \end{table}

\noindent {\bf Results on Something-Something}. Table~\ref{tab:something2_results} compares our CSNs with state-of-the-art methods on Something$^2$-v1 validation set. Our ir-CSN-152, even when trained from scratch, outperforms all of previous methods. Our ip-CSN-152, when pretrained on IG-65M~\cite{uru_video}, achieves $53.3\%$ top-1 accuracy, the new state-of-the-art record for this benchmark. On the same pretraining and finetuning setting, our ir-CSN-152 and ip-CSN-152 outperform R(2+1)D-152 by $0.5\%$ and $1.7\%$, respectively.
 
 \begin{table}
\begin{center}
{\small
\begin{tabular}{|c|c|c|}
\hline 
{\bf Method} & {\bf pretrain} & {\bf video@1} \\ 
\hline 
M-TRN~\cite{abs08496} & ImageNet & 34.4 \\ 
NL I3D~\cite{XiaolongWang18} & ImageNet & 44.4  \\ 
NL I3D + GCN~\cite{XiaolongWangGCN18} & ImageNet & 46.1 \\ 
Motion Feature Net~\cite{MotionNet} & none & 43.9 \\
TSM~\cite{abs-1811-08383} & Kinetics & 44.8 \\
TSM (ensemble)~\cite{abs-1811-08383} & Kinetics & 46.8 \\
ECO-Net~\cite{econet} & ImageNet & 46.4 \\
S3D-G~\cite{xie2017rethinking} & ImageNet & 48.2 \\
\hline
ir-CSN-101 & none & 48.4 \\ 
ir-CSN-152 & none & {\bf 49.3} \\ 
\hline 
R(2+1)D-152*~\cite{uru_video} & IG-65M & 51.6\\ 
ir-CSN-152* & IG-65M & {\bf 52.1} \\
ip-CSN-152* & IG-65M & {\bf 53.3} \\
 \hline
\end{tabular} 
}
\vspace{2pt}
\caption{{\bf Comparisons with state-of-the-art methods on Something$^2$-V1}. Our CSNs outperform all previous methods by good margins when restricting all models to use only RGB as input.}
\label{tab:something2_results}
\end{center}
\vspace{-8pt}
\end{table}

%% file: conclusion.tex
\section{Conclusion}
\label{sec:conclusion}

We have presented Channel-Separated Convolutional Networks (CSN) as a way of factorizing 3D convolutions. The proposed CSN-based factorization not only helps to significantly reduce the computational cost, but also improves the accuracy when there are enough channel interactions in the networks. Our proposed architecture, ir- and ip-CSN, significantly outperform existing methods and obtains state-of-the-art accuracy on three major benchmarks: Sports1M, Kinetics, and Something-Something. The model is also multiple times faster than current competing networks. We have made code and pre-trained models publicly available~\cite{VMZ}.

\noindent {\bf Acknowledgements}. We thank Kaiming He for insightful discussions and Haoqi Fan for help in improving our training framework.

%% file: appendix.tex
\subsection*{Appendix: Visualization of CSN filters}

\begin{figure*}
\begin{center}
   \includegraphics[width=\linewidth]{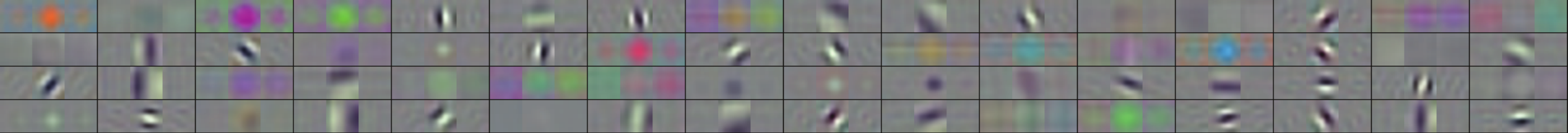}
\end{center}
\vspace{-6pt}
   \caption{Visualization of ir-CSN-152 \texttt{conv\_1} filters. The layer includes 64 convolutional filters, each one of size 3$\times$7$\times$7. Filters are scaled-up by x5 for better presentation. Best viewed in color.}
\label{fig:csn_conv1}
\vspace{-6pt}
\end{figure*}

\begin{figure*}
\begin{center}
   \includegraphics[width=\linewidth]{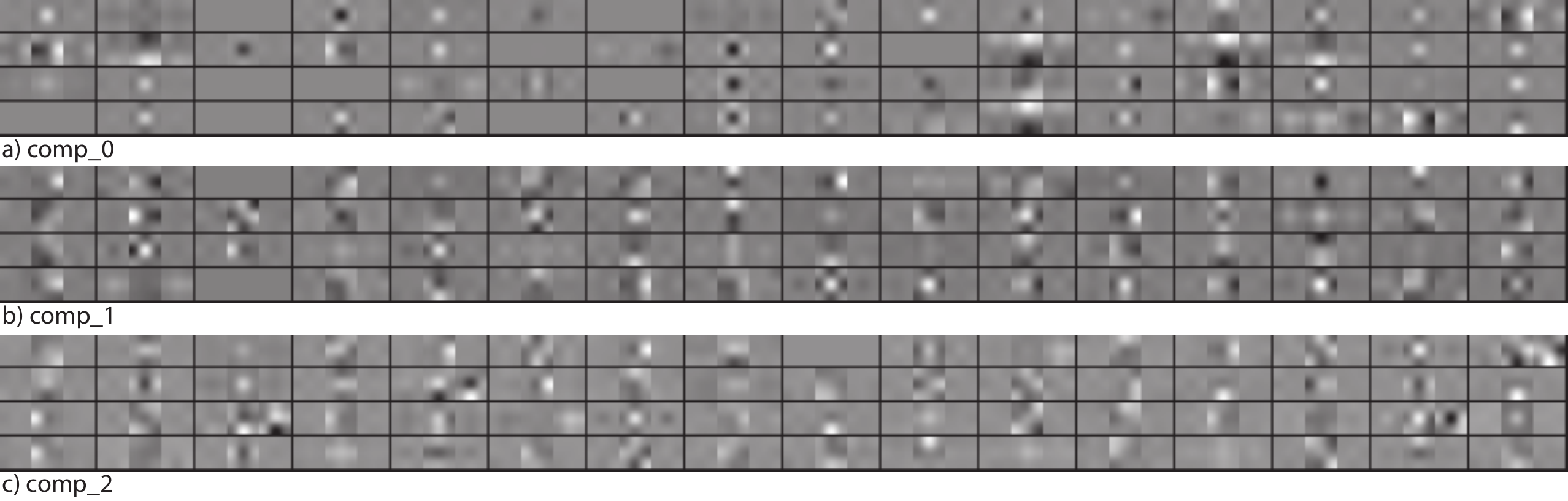}
\end{center}
\vspace{-6pt}
   \caption{Visualization of the 3$\times$3$\times$3 channel-separated convolutional filters in the first three blocks of ir-CSN-152  after \texttt{conv\_1} (and \texttt{pool\_1}): (a) \texttt{comp\_0}, (b) \texttt{comp\_1}, and (c) \texttt{comp\_2}.}
\label{fig:csn_conv2}
\vspace{-6pt}
\end{figure*}

\begin{figure*}
\begin{center}
   \includegraphics[width=\linewidth]{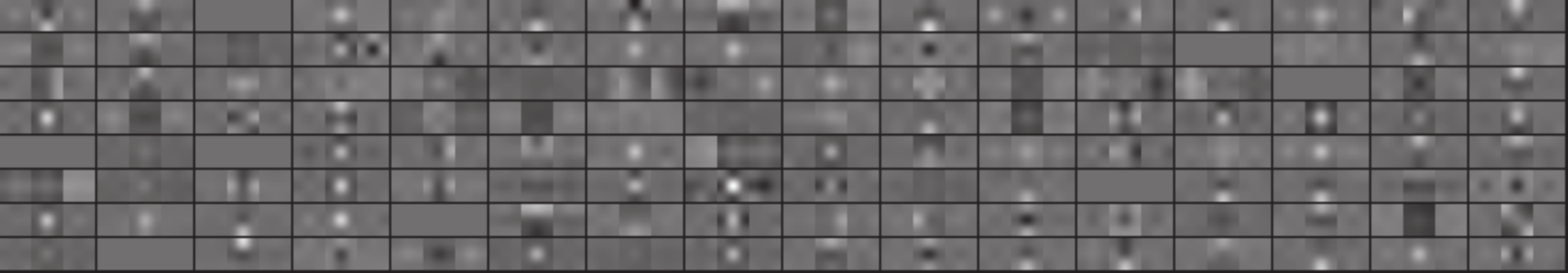}
\end{center}
\vspace{-6pt}
   \caption{Visualization of the 128 3$\times$3$\times$3 channel-separated convolutional filters in \texttt{comp\_10} of ir-CSN-152. \texttt{comp\_10} is an arbitrarily chosen block in the second convolutional group.}
\label{fig:csn_conv3}
\vspace{-6pt}
\end{figure*}

\begin{figure*}
\begin{center}
   \includegraphics[width=\linewidth]{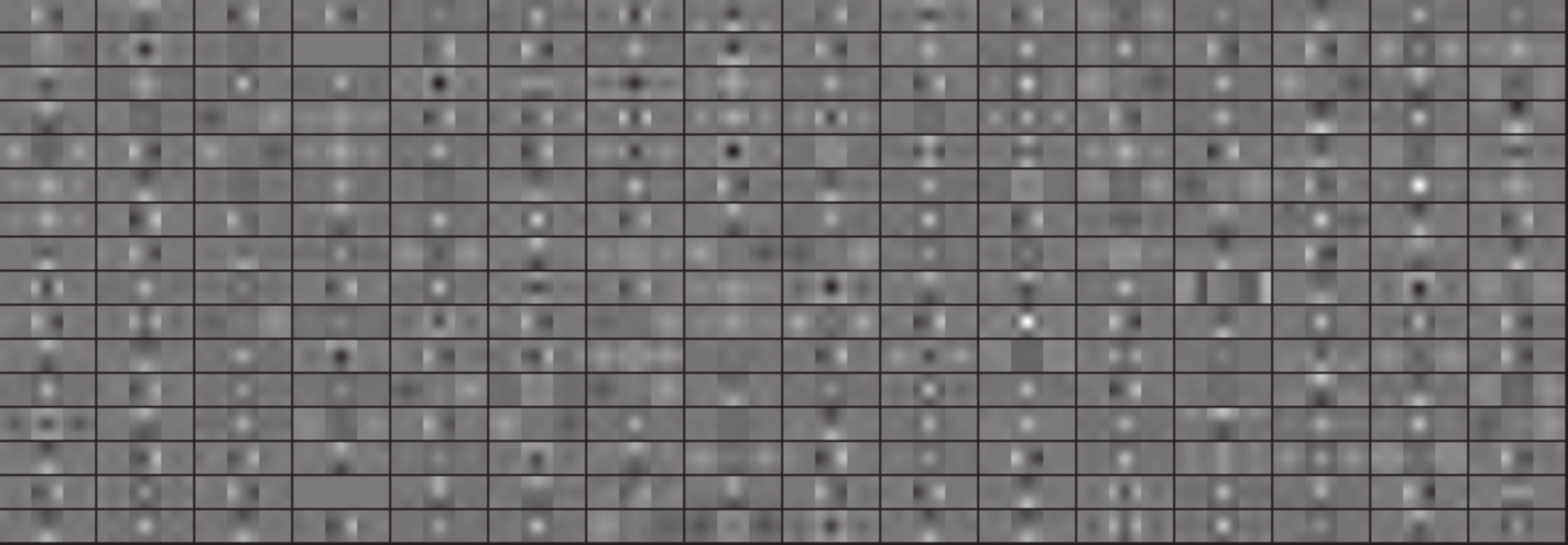}
\end{center}
\vspace{-6pt}
   \caption{Visualization of the 256 3$\times$3$\times$3 channel-separated convolutional filters in \texttt{comp\_12} of ir-CSN-152. \texttt{comp\_12} is an arbitrarily chosen block in the third convolutional group.}
\label{fig:csn_conv4}
\vspace{-6pt}
\end{figure*}

Because all 3$\times$3$\times$3 convolution layers in CSN (except for \texttt{conv\_1}) are depthwise, it is easy to visualize these learned filters. Figure~\ref{fig:csn_conv1}--\ref{fig:csn_conv4} present some of the filters learned by our ir-CSN-152. Our ir-CSN-152 has the first convolution layer, i.e. \texttt{conv\_1} is a normal 3D convolution and 50 bottleneck-D convolutional blocks. These blocks are grouped into four groups by the output size (see Table~\ref{tab:basic_architecture}). The ir-CSN-152 \texttt{conv\_1} filters are presented in Figure~\ref{fig:csn_conv1}. The learned 3$\times$3$\times$3 depthwise convolution filters of the first 3 blocks (named \texttt{comp\_0}, \texttt{comp\_1}, and \texttt{comp\_2}) are presented Figure~\ref{fig:csn_conv2}. Figure~\ref{fig:csn_conv3} and ~\ref{fig:csn_conv4} present the learned 3$\times$3$\times$3 depthwise convolution filters in a random chosen blocks in the 2nd and 3rd convolutional group, respectively. It is interesting to observe that \texttt{conv\_1} captures texture, motion, color changing pattern since it has all cross-channel filters. The depthwise convolution filters learn the motion and/or texture patterns within the same feature channel given by its previous layer. Full images and GIF animations  can be downloaded at \url{https://bit.ly/2KuEsKY}.

